\documentclass{article}

\usepackage[utf8]{inputenc} % allow utf-8 input
\usepackage[T1]{fontenc}    % use 8-bit T1 fonts
\usepackage{hyperref}       % hyperlinks
\usepackage{url}            % simple URL typesetting
\usepackage{booktabs}       % professional-quality tables
\usepackage{amsfonts}       % blackboard math symbols
\usepackage{nicefrac}       % compact symbols for 1/2, etc.
\usepackage{microtype}      % microtypography
\usepackage{amssymb} %maths
\usepackage{amsmath} %mathsx
\usepackage{graphicx}
\usepackage{adjustbox}
\usepackage{algorithm}
\usepackage{multirow}
\usepackage{afterpage}
\usepackage[noend]{algpseudocode}
\usepackage{balance}

\newcommand{\mbf}[1]{\mathbf{#1}}
\newcommand{\cut}[1]{}

\DeclareMathAlphabet{\pazocal}{OMS}{zplm}{m}{n}
\graphicspath{ {./images/} }
\usepackage{subcaption}
\usepackage{color, colortbl, xcolor}

\newcommand{\acro}{{URSABench}}
\newcommand{\name}{{Uncertainty, Robustness, Scalability, and Accuracy Benchmark}}

\usepackage[accepted]{icml2020}

\icmltitlerunning{\acro: Comprehensive Benchmarking of Approximate Bayesian Inference Methods for Deep Neural Networks}

\begin{document}

\twocolumn[
\icmltitle{ \acro: Comprehensive Benchmarking of Approximate\\ Bayesian Inference Methods for Deep Neural Networks}



\icmlsetsymbol{equal}{*}

\begin{icmlauthorlist}
\icmlauthor{Meet P. Vadera}{equal,umass}
\icmlauthor{Adam D. Cobb}{equal,arl}
\icmlauthor{Brian Jalaian}{arl}
\icmlauthor{Benjamin M. Marlin}{umass}
\end{icmlauthorlist}

\icmlaffiliation{umass}{University of Massachusetts Amherst}
\icmlaffiliation{arl}{US Army Research Laboratory}

\icmlcorrespondingauthor{Meet P. Vadera}{mvadera@cs.umass.edu}
\icmlcorrespondingauthor{Adam D. Cobb}{cobb.derek.adam@gmail.com}

\icmlkeywords{Bayesian Neural Networks, MCMC}

\vskip 0.3in
]



\printAffiliationsAndNotice{\icmlEqualContribution} % otherwise use the standard text.
%!TEX root = main.tex
\begin{abstract}
While deep learning methods continue to improve in predictive accuracy on a wide range of application domains, significant issues remain with other aspects of their performance including their ability to quantify uncertainty and their robustness. Recent advances in approximate Bayesian inference hold significant promise for addressing these concerns, but the computational scalability of these methods can be problematic when applied to large-scale models. In this paper, we describe initial work on the development of \emph{\acro} (the \name), an open-source suite of benchmarking tools for  comprehensive assessment of approximate Bayesian inference methods with a focus on deep learning-based classification tasks.\footnote{Our PyTorch implementation is available at \url{https://github.com/reml-lab/URSABench}.} 
% \emph{URSA} stands for \textbf{u}ncertainty, \textbf{r}obustness, \textbf{s}peed, and \textbf{a}ccuracy , which form the basis of our evaluations in this benchmark.
\end{abstract}
%!TEX root = main.tex
\section{Introduction}
\label{sec:intro}

As deep learning models continue to improve their predictive accuracy across many application domains, significant issues remain with respect to other highly important aspects of performance including their ability to robustly quantify uncertainty \cite{guo2017calibration} and their ability to provide robust predictions in the presence of adversarial manipulations \cite{Goodfellow2015ICLR} and out-of-distribution examples \cite{Ovadia2019can}. 

Approximate Bayesian inference methods \cite{Neal:1996:BLN:525544, jaakkola2000bayesian} hold considerable promise for addressing such issues, and recent advances have significantly improved the feasibility of deploying approximate Bayesian inference methods to increasingly larger deep learning models \cite{welling2011bayesian, Zhang2020Cyclical}. 

This paper describes initial work on {\acro}, an open source suite of benchmarking tools for assessment of approximate Bayesian inference methods applied to deep neural network classification tasks. {\acro} includes benchmark models, data sets, tasks and evaluation metrics focused on simultaneously assessing the uncertainty quantification performance, robustness, computational scalability and accuracy of learning and inference methods. We begin by briefly reviewing approximate Bayesian supervised learning. We then discuss principles for evaluation of such methods, followed by a description of the initial {\acro} infrastructure and initial benchmarking results. 

%We also provide a set of benchmark results that for the first time compare an array of recent methods across an array of tasks and evaluation metrics. We hope that the open-source benchmarking infrastructure that we present will be a highly useful tool for the research community and that it will hope to expose areas with the greatest performance gaps to help focus research efforts.

%!TEX root = main.tex
\section{Bayesian Supervised Learning}
\label{sec:related}

In supervised learning, the data set $\mathcal{D}$ consists of a set of labeled instances $\{(\mathbf{x}_i,y_i)|1\leq i\leq N\}$. $\mathbf{x}_i\in\mathbb{R}^D$ is the feature vector and $y_i\in\mathcal{Y}$ is the prediction target. We let $\mathcal{D}^x$ be the set of feature vectors and $\mathcal{D}^y$ be the set of targets.
A probabilistic supervised learning model provides a conditional probability model of the form $p(y|\mathbf{x}, \theta)$ where $\theta\in\mathbb{R}^K$ are the model parameters. The conditional likelihood of the targets given the feature vectors and parameters is given by $p(\mathcal{D}^y|\mathcal{D}^x,\theta)$. The standard assumption that the data cases are independent and identically distributed leads to  $p(\mathcal{D}^y|\mathcal{D}^x,\theta)=\prod_{i=1}^N p(y_i|\mathbf{x}_i, \theta)$ \cite{Neal:1996:BLN:525544}.  Bayesian inference also requires asserting a prior distribution over the model parameters $p(\theta|\theta^0)$ that itself depends on prior parameters $\theta^0$ \cite{Neal:1996:BLN:525544}. 

The two key problems in Bayesian inference applied to supervised learning are the computation of the posterior distribution over the unknown parameters given a training data set $\mathcal{D}_{tr}$ as shown in Equation \eqref{eq:posterior}, and the computation of the posterior predictive distribution over the target variable $y$ given a feature vector $\mbf{x}$ and a data set $\mathcal{D}_{tr}$ as shown in Equation \eqref{eq:posterior_predictive} \cite{Neal:1996:BLN:525544}.
\begin{align}
\label{eq:posterior}
    p(\theta|\mathcal{D}_{tr},\theta^0)& = \frac{p(\mathcal{D}_{tr}^y|\mathcal{D}_{tr}^x,\theta)p(\theta|\theta^0)}{\int p(\mathcal{D}_{tr}^y|\mathcal{D}_{tr}^x,\theta)p(\theta|\theta^0) d\theta}\\
\label{eq:posterior_predictive}
p(y| \mathbf{x}, \mathcal{D}_{tr},\theta^0) &= 
\mathbb{E}_{p(\theta|\mathcal{D}_{tr},\theta^0)}[p(y|\mathbf{x}, \theta)]
\end{align}
It is well known that for neural network models the integrals required in Equations \eqref{eq:posterior} and \eqref{eq:posterior_predictive} are intractable. Approximate Bayesian inference methods thus aim to approximate either the parameter posterior or expectations taken with respect to the parameter posterior such as Equation \eqref{eq:posterior_predictive}. 
Below, we briefly review three  categories of Approximate Bayesian inference methods: Monte Carlo methods, surrogate density methods, and posterior distillation methods.

\textbf{Monte Carlo Methods:} Monte Carlo methods are a classical approach to Bayesian computation that approximate the intractable parameter posterior $p(\theta|\mathcal{D},\theta^0)$ via a distribution constructed from a finite set of samples $\theta_s$ drawn from the true posterior $p(\theta|\mathcal{D},\theta^0)$ \cite{smith1993bayesian}. This leads to the following approximate posterior predictive distribution: $p(y| \mathbf{x}, \mathcal{D},\theta^0)
\approx\frac{1}{S}\sum_{s=1}^S p(y|\mathbf{x}, \theta_s)$.

%shown in Equation \ref{eq:mc_posterior_predictive}.
%
%\begin{align}
%\label{eq:mc_posterior_predictive}
%p(y| \mathbf{x}, \mathcal{D},\theta^0)
%&\approx\frac{1}{S}\sum_{s=1}^S %p(y|\mathbf{x}, \theta_s) 
%\end{align}
%
Of course, for complex models the problem of drawing samples from the true parameter posterior is also often computationally intractable. Markov chain Monte Carlo (MCMC) methods solve this problem by constructing a Markov chain with the true posterior $p(\theta|\mathcal{D},\theta^0)$ as its equilibrium distribution. While classical MCMC methods are typically to slow to apply to large models \cite{casella1992explaining,chib1995understanding,duane1987hybrid,neal2003slice,girolami2011riemann}, a number of recent approaches have addressed this problem either by enabling sampling based on mini-batches of data \cite{welling2011bayesian, Chen2014StochasticGH,Zhang2020Cyclical}, or by sampling in reduced-dimensional parameter spaces \cite{Izmailov2019SubspaceIF}.

\textbf{Surrogate Density Methods:} Another major family of methods are approaches based on approximating the true posterior density via an analytically tractable surrogate distribution $q(\theta|\mathcal{D},\theta^0,\phi)$ where $\phi$ are auxiliary parameters of the surrogate distribution \cite{jordan1999introduction, jaakkola2000bayesian, ghosh2016assumed, minka2001expectation}. 
The most commonly used approaches in this family learn the parameters $\phi$ by minimizing the Kullback-Leibler (KL) divergence $ \textrm{KL}(p||p') = E_{p}[\log (p/p')]$ \cite{mackay2003information}. When the surrogate posterior is used as the first argument, the result is the variational 
inference (VI) framework \cite{jaakkola2000bayesian}. When it is  used as the second argument, it yields the expectation propagation (EP) framework \cite{minka2001expectation}. Advances in the past decade have led to significantly more scalable methods in this family \cite{hoffman2013stochastic,  gal2016dropout, Dusenberry2020EfficientAS}. 

\textbf{Distillation-Based Methods:} The final class of methods that we review are posterior distillation-based methods including Bayesian Dark Knowledge (BDK) \cite{balan2015bayesian} 
and Generalized Posterior Expectation Distillation (GPED) \cite{Vadera2020GeneralizedBP}. 
These methods directly approximate statistics of the posterior distribution by learning an auxiliary neural network model to mimic the output of corresponding Monte Carlo approximations. Importantly, their goal is not to improve over the Monte Carlo approximation, but rather to reduce the computation time required to compute the Monte Carlo average at deployment time. 
%The GPED framework helps identify and mitigate the pitfalls of BDK framework around student architecture search, and also extends it to a general framework for distilling arbitrary posterior expectation statistics.

%\textbf{Discussion}: While Monte Carlo methods provide an unbiased approximation to the posterior, they can be often expensive in terms of computational time required. Surrogate density methods are relatively less expensive in terms of computation time, but they provide a biased estimate to the posterior. Both of these classes  of methods can require substantial time to generate an ensemble to approximate the posterior, and that is the exact problem that distillation methods address. 

% \textbf{Discussion:}  Approaches to approximate Bayesian inference can differ substantially in a number of respects. In principle, MCMC methods are unbiased, but they rely on convergence of Markov chains to produce samples from the posterior. Recent methods have used different acceleration techniques, but can still be significantly slower than SGD-based training, and the resulting posterior ensembles are both storage and compute intensive at test time.  Surrogate density methods have essentially complementary training-time properties. They have a fixed bias determined by the approximating family and discrepancy function used, but can be much faster. At test time, an ensemble is still required, which can make these methods slow as well. Distillation methods do not address training time issues, but can significantly reduce both storage and computation at test time. 

\begin{table*}[!htbp]
\small
  \caption{{\acro} \textbf{small-scale} benchmark performance. Results presented as mean $\pm$ std. dev. across 5 trials.)}
  \label{tab:summary_small}
  \centering
\resizebox{0.8\linewidth}{!}{
\begin{tabular}{@{}cccccr@{}}
\toprule
Inference & Accuracy $\uparrow$ & NLL $\downarrow$  & Robustness $\uparrow$ & Uncertainty $\uparrow$  & Scalability $\downarrow$\\ \midrule
HMC     & 0.9819 $\pm$ 0.0010  & 0.0593 $\pm$ 0.0016 & 0.9570 $\pm$ 0.0075 & 0.9734 $\pm$ 0.0012 & 0.72 $\pm$ 0.01      \\

SGLD      & 0.9839 $\pm$ 0.0004  & 0.0492 $\pm$ 0.0022 & 0.9065 $\pm$ 0.0377 & 0.9679 $\pm$ 0.0233   & 2.02 $\pm$ 0.02      \\

SGHMC     & 0.9862 $\pm$ 0.0003  & 0.0446 $\pm$ 0.0003 & 0.9426 $\pm$ 0.0048 & 0.9807 $\pm$ 0.0003   & 2.03 $\pm$ 0.02     \\

cSGLD    & 0.9857 $\pm$ 0.0003  & 0.0476 $\pm$ 0.0011 & 0.9521 $\pm$ 0.0022 & 0.9795 $\pm$ 0.0007 & 14.08 $\pm$ 0.05      \\

cSGHMC    & 0.9836 $\pm$ 0.0009  & 0.0533 $\pm$ 0.0016 & 0.9276 $\pm$ 0.0094 & 0.9759 $\pm$ 0.0015 &  14.77 $\pm$ 0.03      \\

PCA + ESS (SI)    & 0.9840 $\pm$ 0.0007  & 0.0520 $\pm$ 0.0016 & 0.9360 $\pm$ 0.0038 & 0.9695 $\pm$ 0.0012  & 70.67 $\pm$ 0.20     \\

MC dropout    & 0.9858 $\pm$ 0.0007  & 0.0501 $\pm$ 0.0031 & 0.9429 $\pm$ 0.0059 & 0.9769 $\pm$ 0.0019  & 2.02 $\pm$ 0.03      \\

SGD    & 0.9860 $\pm$ 0.0002  & 0.0452 $\pm$ 0.0012 & - & - &  2.03 $\pm$ 0.02       \\

\bottomrule
        %   &          &       &       &       &               \\ \bottomrule
\end{tabular}
}
\end{table*}

%%%%%%%%%%%%%%%%%%%%%%%%%%%%%%%%%%%%
% Med-scale
%%%%%%%%%%%%%%%%%%%%%%%%%%%%%%%%%%%

\begin{table*}[!htbp]
\small
\caption{{\acro} \textbf{medium-scale} benchmark performance.}
\label{tab:summary-medium}
\centering
\resizebox{0.6\linewidth}{!}{
\begin{tabular}{@{}cccccr@{}}
\toprule
Inference & Accuracy $\uparrow$ & NLL $\downarrow$  & Robustness $\uparrow$ & Uncertainty $\uparrow$  & Scalability $\downarrow$\\ \midrule
SGLD  & 0.869 & 0.524 & 0.803 & 0.916 & 129.3\\ 
SGHMC  & 0.868 & 0.539 & 0.808 & 0.916 & 129.3\\ 
cSGLD  & 0.892 & 0.396 & 0.810 & 0.912 & 2103.3\\ 
cSGHMC  & 0.886 & 0.443 & 0.798 & 0.898 & 2114.9\\ 
SWAG  & 0.824 & 0.735 & 0.759 & 0.885 & 1351.7\\ 
PCA + ESS (SI)  & 0.869 & 0.482 & 0.804 & 0.901 & 1940.0 \\ 
MC dropout  & 0.872 & 0.554 & 0.775 & 0.914 & 127.6 \\ 
% SGD  & 0.861 & 0.625 & 0.643 & 0.772 & 127.7\\ 
SGD  & 0.861 & 0.625 & - & - & 127.7\\ 
\bottomrule
\end{tabular}
}
\end{table*}

\section{{\acro} Evaluation Principles}\label{sec:eval_princ}

While advances in supervised deep learning methods have focused heavily on accuracy over the last decade, there are multiple aspects of models and inference algorithms that are of great interest. {\acro} focuses on simultaneously assessing the uncertainty quantification performance, robustness, computational scalability and accuracy of learning and inference methods. In this section, we describe the evaluation principles that underlie {\acro}. In the next section, we describe their current implementation.

%including their ability to robustly quantify uncertainty and provide correct output in the presence of adversarial manipulations and out of distribution examples. The use of approximate Bayesian inference instead of more common gradient-based point estimation methods holds significant promise for improving performance with respect to these aspects of model performance. 

\textbf{Accuracy:} Predictive performance is by far the most widely considered property of supervised machine learning models. In the classification setting, evaluation measures that only require that the true label $y$ be correctly predicted provide the coarsest measures of the predictive performance. Accuracy is the most common such measure.
% Such measures include accuracy, precision, recall, and F1 score. 

%In the classification setting, predictive log likelihood can provide a more refined notion of predictive performance by assessing the log probability of the true class. This measure is strongly influenced by confident incorrect predictions. In the regression case, the use of predictive log likelihoods for evaluation can also provide more information about the posterior uncertainty of the predictions. However, one of the main drawbacks of predictive log likelihood is that absolute log likelihood values are not as easily interpretable as absolute accuracy values. Predictive likelihood is often better suited for use in relative comparisons between models or inference methods as a result.

%Measures like AUC are sensitive to predictive probabilities and can provide more information about the quality of probabilistic predictions for binary classification problems.

\textbf {Uncertainty Quantification:} A number of metrics are helpful for assessing the degree to which a method results in properly quantified predictive uncertainty.  Predictive log likelihood can provide more insight into the predictive distribution than accuracy as it is sensitive to the predicted value of $p(y| \mathbf{x}, \mathcal{D}_{tr},\theta^0)$. Both high-confidence errors and low-confidence correct predictions will result in lower log likelihood values. 

Calibration is also an important property of predictive models and recent evaluations of deep learning methods have shown that their calibration properties can be quite poor \cite{guo2017calibration}. 
In the binary case, a classifier is said to be perfectly calibrated if exactly $p$ percent of instances are predicted to be positive with $p$ percent probability. 
%The expected calibration error of a predictive model can be estimated by binning the predictive probabilities in $N_b$ bins and computing the accuracy $A_b$ and average confidence $C_b$ within each bin $b$. The calibration error for bin $b$ is defined to be $\kappa_b=|A_b-C_b|$ and the expected calibration error (ECE) is then defined as $ECE=\frac{1}{N}\sum_b N_b\kappa_b$ \cite{guo2017calibration}. 
The degree of calibration of a binary classifier can be quantified using the expected calibration error (ECE) \cite{guo2017calibration}. In the case of multi-class classification, a one-vs-all formulation of calibration error can be used. The Brier score provides an alternate measure of calibration \cite{brier1950verification} that can be interpreted as mixing together aspects of calibration and accuracy. Finally, misclassification detection performance \cite{Hendrycks2017ABF} is also helpful in assessing the utility of various uncertainty metrics.
%Like prediction performance measures, calibration is assessed on an in-domain test data set $\mathcal{D}_{te}$ that is assumed to be sampled from the same distribution as the training data set $\mathcal{D}_{te}$. 

\textbf{Robustness:} Another key property of models and inference methods is their robustness. Both predictive performance and uncertainty quantification metrics are typically computed on a test data set $\mathcal{D}_{te}$ that is assumed to be sampled from the same distribution as the training data set $\mathcal{D}_{tr}$. Out-of-distribution (OOD) detection tasks instead assess the ability of methods to detect examples from a set $\mathcal{D}_{ood}$ drawn from a different distribution than $\mathcal{D}_{tr}$ \cite{Ovadia2019can}. The ability of methods to resist adversarial input perturbations as measured by the success rate of different adversarial attacks is also an important property \cite{Goodfellow2015ICLR, madry2018towards,Carlini2017}. We note that in the Bayesian supervised learning context, these attack methods require access to the posterior predictive distribution function and in many cases its gradients \cite{Vadera2020AssessingTA}.       

\textbf{Scalability:} Of primary interest in this work are how the accuracy, uncertainty quantification and robustness properties of methods trade off against their computational scalability properties with the goal of better understanding which methods offer the best trade-offs in different computational contexts (e.g., cluster, embedded system, etc.). The storage cost can be estimated via the number of parameters and the size of stored models (if variable bit depth is considered). The run-time of methods can be assessed in different ways including wall clock time as well as more portable statistics such as the number of floating point operations (flops) or  multiply-accumulate operations (MACs).

\section{{\acro} Implementation Framework} \label{sec:framework}

In this section, we describe the current {\acro} implementation framework, which leverages multiple datasets, models and tasks to implement the evaluation principles described in the previous section. The current framework includes small-scale and medium-scale benchmarks. 

\textbf{Models and Data Sets: } The small-scale benchmark uses a basic, fully connected MLP with two hidden layers containing $200$ units each as the benchmark model, with  MNIST  providing the benchmark in-domain data set \cite{lecun1998mnist}. At the medium-scale, we use ResNet50 \cite{he2016deep} and WideResNet  as the benchmark models \cite{Zagoruyko2016WideRN}, with CIFAR10 and CIFAR100 as the benchmark in-domain data sets \cite{krizhevsky2009learning}.

\textbf{Tasks and Metrics: } The approximate parameter posterior and posterior predictive distribution are produced using each benchmark in-domain training set. Accuracy is assessed using the corresponding in-domain test sets. To assess uncertainty quantification, we compute negative log likelihood, Brier score, and performance on a misclassification detection task, all using the in-domain test sets. We also consider a decision-making task that focuses on assessing the quality of the tail of the predictive distribution using imbalanced data sets and costs that strongly penalize errors on the rare classes \cite{cobb2018loss} (see Appendix \ref{ap:decision_making} for details). We assess robustness using an out-of-distribution (OOD) classification task \citep{Ovadia2019can, Vadera2020GeneralizedBP} leveraging knowledge uncertainty (see Appendix \ref{ap:uncertainty_decomposition} for a review of uncertinaty decomposition). The small-scale benchmark uses FashionMNIST \cite{Xiao2017FashionMNISTAN} and KMNIST \cite{clanuwat2018kmnist} as OOD test sets, while the medium-scale benchmark uses SVHN \cite{Netzer2011SVHN} and STL10 \cite{Coates2011STL10} as OOD test sets. Performance on OOD tasks is assessed using AUROC. Finally, the current version of the benchmark focuses on computation time as the measure of computational scalability, measured in seconds/sample.

\textbf{Composite Scores:} The simultaneous assessment of multiple aspects of performance is the focus of {\acro}. However, this yields many individual results for each inference method. An important design choice in {\acro} is thus to summarize performance in terms of key selected individual metrics along with composite scores that combine related individual metrics. For accuracy we include an average over all benchmark models and all in-domain test sets. For robustness, we use an average over all models and OOD data sets. For uncertainty quantification, we separately compute an average over models and data sets in terms of negative log likelihood (NLL) and misclassification task performance.

\section{{\acro} Benchmark Results} 

In this section we report the initial benchmark results obtained using {\acro}.

\textbf{Inference Methods: } We focus on bencmarking Monte Carlo methods including HMC\footnote{HMC is only implemented for tasks where the model and full data set can fit on the GPU. We use the \texttt{hamiltorch} Python package \cite{cobb2019introducing}.}, SGLD \cite{welling2011bayesian}, SGHMC \cite{Chen2014StochasticGH}, cSGLD, cSGHMC \cite{Zhang2020Cyclical}, SWAG \cite{Maddox2019ASB} and PCA-based subspace inference + elliptical slice sampling (PCA+ ESS (SI)) \cite{Izmailov2019SubspaceIF}. As baselines, we also provide MC dropout \cite{gal2016dropout} and an SGD-point estimated model. Implementation details for the inference schemes have been provided in Appendix \ref{ap:implementation_details}.

\textbf{Small-Scale Benchmark Results:} The small-scale results are displayed in Table \ref{tab:summary_small}. The detailed experimental results behind each composite score can be found in Appendix \ref{ap:results}. The small-scale results show how challenging it can be to distinguish between  different approximate inference schemes using relatively simple models and data sets. SGD and SGHMC are both marginally ahead in accuracy and NLL; HMC appears to show the most robust performance in OOD and SGHMC does best for the uncertainty metric. However the minor relative difference between all the performance metrics points to focusing on the compute time which shows HMC to be significantly less time consuming. This is due to the ability to fit all the data and model parameters on the GPU. 

\textbf{Medium-Scale Benchmark Results:} The medium-scale results are displayed in Table \ref{tab:summary-medium}. The detailed experimental results behind each composite score can again be found in Appendix \ref{ap:results}. Overall, the medium-scale experiments indicate a slight improvement on the predictive performance and decision-making tasks from both cSGHMC, and cSGLD followed by PCA + ESS (SI). However, once again a user may prefer using MC dropout or SGLD/SGHMC as they provide respectable performance in significantly less time. This is due to the large proportion of time that the cyclic schemes spend exploring without sampling. Furthermore, if the goal is to compute uncertainty metrics and ultimately use them for misclassification detection or OOD detection, then SGHMC/SGLD provide better performance in a majority of the cases.  Another important result that can be seen from the Tables \ref{tab:pred-resnet50-cifar10}, \ref{tab:pred-resnet50-cifar100}, \ref{tab:pred-wide-cifar10} and \ref{tab:pred-wide-cifar100} in Appendix \ref{ap:results} is the demonstrated utility of the decision-making task in its ability to highlight the top performing approximate inference schemes for each model and data set, via its correlation with low NLL and high accuracy.

% \paragraph{Paragraph summarising results}

% \section{Experiments}
% We now move towards evaluating different MCMC inference methods on the tasks mentioned in the previous section. For a benchmark like ours, its important to evaluate on a wide range of models and data sets. Thus, for our paper, we run evaluadtions on multiple data sets such as MNIST \cite{lecun1998mnist}, CIFAR10/100 \cite{krizhevsky2009learning}, and Tiny Imagenet \cite{stanfordTim}. We also consider a wide range of models ranging from a simple multi-layer perceptron to more modern architectures like ResNet \cite{he2016deep}, WideResNet \cite{Zagoruyko2016WideRN}, PreResNet \cite{He2016IdentityMI}, and VGG \cite{simonyan2014very}.
% \label{sec:exp}

    % \begin{table}[ht]
    %     \begin{center}
    %         \begin{tabular}{cc*{2}{Acc}}
    %         \multicolumn{1}{c}{Inference} &\multicolumn{1}{c}{Acc}\\
    %           HMC & 1.00 & 1.0 \\
    %         \end{tabular}
    %     \end{center}
    % \end{table}
%!TEX root = main.tex
\section{Conclusion and Future Work}
\label{sec:conclusion}

This paper describes initial work on {\acro}, a framework for benchmarking the performance of approximate Bayesian inference methods for deep neural networks. We hope that the development of this benchamrking toolbox will help to accelerate research in the domain of approximate Bayesian inference by helping to expose the trade-offs achieved by methods in terms of uncertainty, robustness, scalability and accuracy. We believe the simultaneous assessment of these properties is critical to better understand which methods are most effective on different downstream tasks and in different deployment contexts. 

A further clear challenge in the development of this toolbox is ensuring a fair comparison between approaches. However, this can be difficult for new model/method/data set combinations without established hyperparameters, requiring careful hyperparameter optimization. This requirement highlights the issue of how to benchmark the end-to-end process of hyperparameter optimization and inference in terms of computational resource use. 

As a first line of future work, we plan to continue to implement tasks and metrics to fully reflect all of the evaluation principles described in this paper. Important tasks and metrics yet to be implemented include robustness to adversarial examples \cite{Vadera2020AssessingTA} and common corruptions \cite{hendrycks2018benchmarking}, and assessment of test-time computational scalability. We further plan to add a large-scale benchmark that current approximate inference schemes will find challenging. Finally, we aim to expand the scope of models and data sets to include architectures such as recurrent neural networks and graph convolutional networks to provide a broader assessment of approximate inference methods.

% We overcame this approach for the small-scale experiment by assigning the same resource to all schemes when performing Bayesian optimisation. However incorporating the cost of each individual BO evaluation ought to come into consideration when tuning for hyperparameters. Therefore we envisage future work to couple the cost of hyperparameter optimization with the evaluation of a particular inference approach.
\section*{Acknowledgement}
Research reported in this paper was sponsored in part by the CCDC Army Research Laboratory under Cooperative Agreement W911NF-17-2-0196 (ARL IoBT CRA). The views and conclusions contained in this document are those of the authors and should not be interpreted as representing the official policies, either expressed or implied, of the Army Research Laboratory or the U.S. Government. The U.S. Government is authorized to reproduce and distribute reprints for Government purposes notwithstanding any copyright notation herein.

\bibliography{references}
\bibliographystyle{icml2020}

\appendix

% \section*{Supplementary Material}

\section{Decision-making task}\label{ap:decision_making}

Bayesian decision theory takes Monte Carlo samples and averages them over a predetermined cost function, $\mathcal{C}(h,y)$ to result in an expected risk:
$$R(h \vert \mathbf{x}) \approx \frac{1}{S}\sum_{s=1}^S p(y|\mathbf{x}, \theta_s) \mathcal{C}(h,y).$$
The expected risk is a function of the decision, $h$, whereby the Bayesian optimal decision, $h^*$ minimises the risk: $$h^* = \underset{h}{\mathrm{argmin}}\ R(h \vert \mathbf{x}).$$
Once we have applied Bayesian decision theory to find an $h^*$ for every input $\mathbf{x}$, for supervised classification, we can then determine the true cost of the decision taken by averaging over the test data (i.e. $\frac{1}{N}\sum_{n=1}^N\mathcal{C}(y^{\mathrm{True}}_n,h_n^*)$), where the arguments have been reversed such that we compare the true label $y^{\mathrm{True}}_n$, with the Bayesian optimal decision $h_n^*$ (i.e. what cost did we actually have to pay when we took decision $h_n^*$ for labelling $y_n$, when it was in fact $y^{\mathrm{True}}_n$). A more detailed discussion can be seen in Ch. 4 of \citet{cobb2020practicalities}.

The purpose of the decision-making task is to penalise inference schemes that provide poor calibrated uncertainty over the rarer (and hence more uncertain) classes. In particular, for MNIST, we retrain our models over a highly imbalanced data set, where $99 \%$ of the labels corresponding to classes $3$ and $7$ are removed. However, we then use the predictive distribution to with a predefined cost function to select the Bayes optimal decision for each predicted label. We then calculate the expected decision cost by averaging over the costs attributed to each decision compared to the true label. False negatives of the less frequent classes are penalised $1000$ times more than false positives for the rest of the classes.

The small-scale setting for the decision-making task requires retraining over an imbalanced training set. However, for the medium-scale task we limit ourselves to using the same materialised samples from the balanced data set (although we expect to extend this to imbalanced training data in future work). We define our cost matrix to penalise false negatives $10$ times as much as false positives. In particular, the task for the CIFAR10 penalises planes, automobiles, ships and trucks with a cost of $1.0$ for false negatives and $0.1$ for false positives. All other errors are penalised with $0.1$ and correct decisions accrue zero cost. The same cost structure applies to CIFAR100, where tanks, rockets and pick-up trucks are deemed the critical classes.

The decision costs in Tables \ref{tab:pred-resnet50-cifar10}, \ref{tab:pred-resnet50-cifar100}, \ref{tab:pred-wide-cifar10} and \ref{tab:pred-wide-cifar100} demonstrate the utility of this task as they show a correlation with the NLL and the accuracy across all models and data sets.

\section{Uncertainty Decomposition for downstream tasks} \label{ap:uncertainty_decomposition}
The posterior predictive distribution is not the only statistic of the posterior distribution that is of interest. 
The decomposition of posterior uncertainty has also received recent attention in the literature. For example, \citet{Depeweg2017DecompositionOU} and \citet{malinin2020ensemble} describe the decomposition of the entropy of the posterior predictive distribution (the \emph{total uncertainty}) into \emph{expected data uncertainty} and \emph{knowledge uncertainty}. These three forms of uncertainty are related by the equation shown below:

\begin{align}
\underbrace{\pazocal{I}\left[y, \theta | \boldsymbol{x}, \pazocal{D}\right]}_{\text {Knowledge Uncertainty }} & =\underbrace{\pazocal{H}\left[\mathbb{E}_{p(\theta | \pazocal{D})}\left[p\left(y | \boldsymbol{x}, \theta\right)\right]\right]}_{\text {Total Uncertainty }} \notag \\
&-\underbrace{\mathbb{E}_{p(\theta | \pazocal{D})}\left[\pazocal{H}\left[p\left(y | \boldsymbol{x}, \theta\right)\right]\right]}_{\text {Expected Data Uncertainty }}
\end{align}

Total uncertainty, as the name suggests, measures the total uncertainty in a prediction. Expected data uncertainty measures the uncertainty arising from class overlap.   Knowledge uncertainty corresponds to the conditional mutual information between labels and model parameters and measures the disagreement between different models in the posterior. 
However, it can be efficiently computed as the difference between total uncertainty and expected data uncertainty, both of which are (functions) of posterior expectations. In recent work, \citet{wang2018adversarial}, \citet{malinin2020ensemble} and \citet{Vadera2020GeneralizedBP} have leveraged this decomposition to explore a range of down-stream tasks that rely on uncertainty quantification and decomposition. 

\section{Composite Score Breakdown}\label{ap:scores}
As alluded to in the main text, we build composite scores for robustness and uncertainty. The robustness relies on averaging both the total uncertainty AUROC and the model uncertainty AUROC over the OOD data sets. We then average once again over the mean total uncertainty and model uncertainty. The uncertainty composite score is built from the average misclassification AUROCs (e.g. the first three columns of Tables \ref{tab:misclass-mnist}, \ref{tab:misclass-resnet50-cifar10}, \ref{tab:misclass-resnet50-cifar100}, \ref{tab:misclass-wide-cifar10}, \ref{tab:misclass-wide-cifar100}). For the medium-scale experiment the uncertainty score is then averaged across CIFAR10 and CIFAR100 as well as ResNet50 and WideResNet28x10.

\section{Implementation Details}\label{ap:implementation_details}
In this section, we describe the implementation details for the different inference methods used in our benchmark. It must be noted that for all inference methods using ResNet50 and WideResNet28x10 models, we use a pretrained SGD solution to warm-start our samplers. This is a standard pretraining procedure followed to make the methods more competitive \cite{Maddox2019ASB}. Further, the ensemble size is set to 50 for CIFAR datasets, and 100 for MNIST dataset. The difference in ensemble size is due to the large amounts of computational requirements for training ResNet50 and WideResNet28x10 on CIFAR datasets. While tuning hyperparameters for MNIST, we apply Bayesian optimization with a limit of $200$ evaluations for each approach \cite{balandat2019botorch}. On the other hand, for CIFAR datasets, we refer to existing literature and use the same hyperparameters if directly applicable, or search around the hyperparameters obtained for similar models and datasets.

\textbf{SGLD:} For CIFAR datasets, we use a burn-in of 100 epochs and initial learning rates of 0.1 for WideResNet28x10 model and 0.05 for ResNet50. The prior std. dev. is set to 1 for both the cases. We decay the learning rate using cosine annealing schedule to its half value by the end of sampling. For MNIST, the optimal hyperparameter values obtained are: initial learning rate of 0.099, prior std. dev. of 0.16 and 50 burn in epochs.

\textbf{SGHMC:} We use the same hyperparameters and learning rate schedule as described for SGLD for the CIFAR datasets. Additionally, we set the friction term to 0.5 \cite{Chen2014StochasticGH}. This is equivalent to the $\alpha$ term shown in \citet{Zhang2020Cyclical}. For MNIST, the optimal hyperparameter values obtained are: initial learning rate of 0.03, prior std. dev. of 0.14, 50 burn in epochs, and friction term set to 0.1.

\textbf{cSGHMC:} We use the same hyperparameters given in \citet{Zhang2020Cyclical} for CIFAR datasets. For MNIST, the optimal hyperparameter values obtained are: initial learning rate of 0.06, prior std. dev. of 0.33, cycle length of 22 epochs, of which 17 epochs are used for SGD-exploration phase, and samples are collected from the last 4 epochs, and friction term set to 0.21.

\textbf{cSGLD:} We use the same hyperparameters given in \citet{Zhang2020Cyclical} for CIFAR datasets. For MNIST, the optimal hyperparameter values obtained are: initial learning rate of 0.06, prior std. dev. of 0.33, cycle length of 22 epochs, of which 17 epochs are used for SGD-exploration phase, and samples are collected from the last 4 epochs, and friction term set to 0.21.

\textbf{SWAG:} We use the same hyperparameters given in \citet{Izmailov2019SubspaceIF} for CIFAR models, except that we set the weight decay for ResNet models to $10^{-4}$ and borrow its remaining hyperparameters from WideResNet28x10. This means that we utilize last 20 SGD iterates to find parameters for the gaussian approximation to the mode. For MNIST, we start with an initial learning rate of 0.018 and decay it 0.0006 over 50 epochs. Further, we run SGD at the same learning rate for another 30 epochs and collect the final 20 iterates to construct our SWAG approximation. The momentum for our SGD optimizer is set to 0.7 through the entire run. Furthermore, the variant of SWAG used in our benchmark is SWAG-diagonal.

\textbf{PCA + ESS (SI):} We use the same hyperparameters given in \citet{Izmailov2019SubspaceIF} for CIFAR models, except that we set the weight decay for ResNet models to $10^{-4}$ and borrow its remaining hyperparameters from WideResNet28x10. We construct a subspace of rank 20 for all models and datasets. For MNIST, we start with an initial learning rate of 0.04 and decay it 0.002 over 50 epochs. Further, we run SGD at the same learning rate for another 50 epochs and collect the iterates from each of the final 20 epochs to construct our PCA subspace. The momentum for our SGD optimizer is set to 0.54 through the entire run. For all the dataset and model combinations, we use elliptical slice sampling \citep{Murray2010EllipticalSS} on the low rank PCA subspace with a prior of 2. and a temperature of 5000. 

\textbf{MC Dropout:} For all the models on CIFAR datasets, we use a dropout of 0.2 before the final linear layer while we use dropout after both hidden layers for MNIST-MLP200 with a dropout rate of 0.04.

\section{Additional Experimental Results} \label{ap:results}
Additional experimental results are provided in Tables \ref{tab:predictive-mlp200-mnist} - \ref{tab:misclass-wide-cifar100}.
%%%%%%%%%%%%%%%%%%%%%%%%%%%%%%%%%%%%%%%%%%%%%%%%%%%%%%%%%%%%%%%%%%%%%%%%
%  MNIST Tables
%%%%%%%%%%%%%%%%%%%%%%%%%%%%%%%%%%%%%%%%%%%%%%%%%%%%%%%%%%%%%%%%%%%%%%%%
\begin{table*}[htbp]
  \caption{Comparison of predictive performance and decision making cost while using an MLP $[784,200, 200,10]$ on MNIST. Results presented as mean $\pm$ std. dev. across 5 trials.}
  \label{tab:predictive-mlp200-mnist}
  \centering
\resizebox{\linewidth}{!}{
\begin{tabular}{@{}ccccccc@{}}
\toprule
Inference & Accuracy $\uparrow$ & NLL $\downarrow$  & BS $\downarrow$   & ECE $\downarrow$  & Decision Cost $\downarrow$ & Samples/second $\downarrow$\\ \midrule
HMC     & 98.19 $\pm$ 0.10\%  & 0.0593 $\pm$ 0.0016 & 0.0280 $\pm$ 0.0008 & 0.0079 $\pm$ 0.0008 & 7101 $\pm$ 346  & 0.72 $\pm$ 0.01      \\

SGLD      & 98.39 $\pm$ 0.04\%  & 0.0492 $\pm$ 0.0022 & 0.0236 $\pm$ 0.0005 & 0.0041 $\pm$ 0.0024 & 5410 $\pm$ 778  & 2.02 $\pm$ 0.02      \\

SGHMC     & 98.62 $\pm$ 0.03\%  & 0.0446 $\pm$ 0.0003 & 0.0210 $\pm$ 0.0002 & 0.0073 $\pm$ 0.0004 & 5408 $\pm$ 240   & 2.03 $\pm$ 0.02     \\

cSGLD    & 98.57 $\pm$ 0.03\%  & 0.0476 $\pm$ 0.0011 & 0.0223 $\pm$ 0.0003 & 0.0056 $\pm$ 0.0003 & 6526 $\pm$ 2241  & 14.08 $\pm$ 0.05      \\

cSGHMC    & 98.36 $\pm$ 0.09\%  & 0.0533 $\pm$ 0.0016 & 0.0256 $\pm$ 0.0010 & 0.0033 $\pm$ 0.0003 & 4824 $\pm$ 1855  & 14.77 $\pm$ 0.03      \\

PCA + ESS (SI)    & 98.40 $\pm$ 0.07\%  & 0.0520 $\pm$ 0.0016 & 0.0251 $\pm$ 0.0007 & 0.0036 $\pm$ 0.0005 & 3809 $\pm$ 1150  & 70.67 $\pm$ 0.20  \\

MC dropout    & 98.58 $\pm$ 0.07\%  & 0.0501 $\pm$ 0.0031 & 0.0218 $\pm$ 0.0008 & 0.0042 $\pm$ 0.0006 & 15236 $\pm$ 1184  & 2.02 $\pm$ 0.07      \\

SGD    & 98.60 $\pm$ 0.02\%  & 0.0452 $\pm$ 0.0012 & 0.0213 $\pm$ 0.0003 & 0.0032 $\pm$ 0.0005 & 8613 $\pm$ 1428  & 2.03 $\pm$ 0.02      \\ \bottomrule
\end{tabular}
}
\end{table*}

\begin{table*}[]
\caption{Comparison of OOD detection performance while using an MLP $[784,200, 200,10]$ on MNIST. Results presented as mean $\pm$ std. dev. across 5 trials.}
  \label{tab:ood-mlp200-mnist}
  \centering
% \resizebox{\linewidth}{!}{
\begin{tabular}{@{}cccc@{}}
\toprule
Inference &
  \begin{tabular}[c]{@{}c@{}}OOD\\ Dataset\end{tabular} &
  \begin{tabular}[c]{@{}c@{}}AUROC- Model\\ Uncertainty $\uparrow$ \end{tabular}  &
  \begin{tabular}[c]{@{}c@{}}AUROC - Total\\ Uncertainty $\uparrow$ \end{tabular}  \\ \midrule
\multirow{2}{*}{HMC} & Fashion MNIST & 0.966 $\pm$ 0.013 & 0.946 $\pm$ 0.017 \\ %\cline{2-4}
                       & KMNIST  & 0.968 $\pm$ 0.013 & 0.948 $\pm$ 0.017 \\

\multirow{2}{*}{SGLD} & Fashion MNIST & 0.867 $\pm$ 0.110 & 0.944 $\pm$ 0.005 \\ %\cline{2-4}
                       & KMNIST  & 0.871 $\pm$ 0.103 & 0.944 $\pm$ 0.005 \\

\multirow{2}{*}{SGHMC} & Fashion MNIST & 0.933 $\pm$ 0.009 & 0.953 $\pm$ 0.010 \\ %\cline{2-4}
                       & KMNIST  & 0.932 $\pm$ 0.009 & 0.952 $\pm$ 0.010 \\
                       
\multirow{2}{*}{cSGLD} & Fashion MNIST & 0.954 $\pm$ 0.004 & 0.950 $\pm$ 0.005 \\ %\cline{2-4}
                       & KMNIST  & 0.954 $\pm$ 0.004 & 0.950 $\pm$ 0.005 \\
                       
\multirow{2}{*}{cSGHMC} & Fashion MNIST & 0.923 $\pm$ 0.021 & 0.931 $\pm$ 0.017 \\ %\cline{2-4}
                       & KMNIST  & 0.923 $\pm$ 0.020 & 0.933 $\pm$ 0.017 \\

\multirow{2}{*}{PCA + ESS (SI)} & Fashion MNIST & 0.933 $\pm$ 0.006 & 0.938 $\pm$ 0.009 \\ %\cline{2-4}
                       & KMNIST  & 0.934 $\pm$ 0.006 & 0.940 $\pm$ 0.009 \\ 

\multirow{2}{*}{MC dropout} & Fashion MNIST & 0.942 $\pm$ 0.013 & 0.943 $\pm$ 0.010 \\ %\cline{2-4}
                       & KMNIST  & 0.943 $\pm$ 0.013 & 0.944 $\pm$ 0.010 \\

\multirow{2}{*}{SGD} & Fashion MNIST & N/A  &  0.945 $\pm$ 0.010 \\
& KMNIST &  N/A  &  0.943 $\pm$ 0.010 \\                      
                       \bottomrule
\end{tabular}
% }
\end{table*}

\begin{table*}[htbp]
  \caption{Comparison of misclassification detection while using an MLP $[784,200, 200,10]$ on MNIST.}
  \label{tab:misclass-mnist}
  \centering
\resizebox{\linewidth}{!}{
\begin{tabular}{@{}cccllll@{}}
\toprule
Inference &
  \begin{tabular}[c]{@{}c@{}}AUROC- Model\\ Uncertainty $\uparrow$ \end{tabular} &
  \begin{tabular}[c]{@{}c@{}}AUROC - Total\\ Uncertainty $\uparrow$ \end{tabular} &
  \begin{tabular}[c]{@{}l@{}}AUROC- Model\\ Confidence $\uparrow$ \end{tabular} &
  \multicolumn{1}{c}{\begin{tabular}[c]{@{}c@{}}AUCPR- Model\\ Uncertainty $\uparrow$ \end{tabular}} &
  \multicolumn{1}{c}{\begin{tabular}[c]{@{}c@{}}AUCPR - Total\\ Uncertainty $\uparrow$ \end{tabular}} &
  \begin{tabular}[c]{@{}l@{}}AUCPR- Model\\ Confidence $\uparrow$ \end{tabular} \\ \midrule
HMC  &  0.9706  &  0.9734  &  0.9743  &  0.3429  &  0.3888  &  0.4145  \\ 
SGLD  &  0.9739  &  0.9800  &  0.9800  &  0.3530  &  0.4502  &  0.4632  \\ 
SGHMC  &  0.9786  &  0.9815  &  0.9823  &  0.3546  &  0.3929  &  0.4131  \\ 
cSGLD  &  0.9769  &  0.9801  &  0.9798  &  0.3695  &  0.4477  &  0.4478  \\ 
cSGHMC  &  0.9730  &  0.9786  &  0.9786  &  0.3260  &  0.4255  &  0.4404  \\ 
PCA + ESS (SI)  &  0.9539  &  0.9774  &  0.9772  &  0.2059  &  0.4298  &  0.4248  \\ 
MC dropout  &  0.9754  &  0.9763  &  0.976  &  0.4085  &  0.43  &  0.4199  \\
SGD  &  N/A  &  0.9795  &  0.9794  &  N/A  &  0.4273  &  0.4389  \\
\bottomrule
\end{tabular}
}
\end{table*}

%%%%%%%%%%%%%%%%%%%%%%%%%%%%%%%%%%%%%%%%%%%%%%%%%%
% RESNET 50: CIFAR10
%%%%%%%%%%%%%%%%%%%%%%%%%%%%%%%%%%%%%%%%%%%%%%%%%%

\begin{table*}[htbp]
  \caption{Comparison of predictive performance and decision making cost
  while using ResNet50 on CIFAR10.}
  \label{tab:pred-resnet50-cifar10}
  \centering
% \resizebox{\linewidth}{!}{
\begin{tabular}{@{}cccccc@{}}
\toprule
Inference & Accuracy $\uparrow$ & NLL $\downarrow$  & BS  $\downarrow$  & ECE $\downarrow$  & Decision Cost $\downarrow$ \\ \midrule
SGLD  & 0.954 & 0.144 & 0.069 & 0.009 & 139.500  \\ 
SGHMC  & 0.954 & 0.144 & 0.068 & 0.011 & 138.100  \\ 
cSGLD  & 0.966 & 0.128 & 0.053 & 0.020 & 112.100  \\ 
cSGHMC  & 0.951 & 0.243 & 0.086 & 0.106 & 153.900  \\ 
SWAG  & 0.931 & 0.311 & 0.114 & 0.047 & 200.900  \\ 
PCA + ESS (SI)  & 0.949 & 0.174 & 0.080 & 0.027 & 166.600  \\ 
MC dropout  & 0.948 & 0.208 & 0.083 & 0.032 & 159.500  \\ 
SGD  & 0.943 & 0.274 & 0.095 & 0.040 & 171.700  \\  \bottomrule
\end{tabular}
% }
\end{table*}

\begin{table*}[htbp]
\caption{Comparison of OOD detection performance while using ResNet50 on CIFAR10.}
  \label{tab:ood-resnet50-cifar10}
  \centering
\begin{tabular}{@{}cccc@{}}
\toprule
Inference &
  \begin{tabular}[c]{@{}c@{}}OOD\\ Dataset\end{tabular} &
  \begin{tabular}[c]{@{}c@{}}AUROC- Model\\ Uncertainty $\uparrow$ \end{tabular}  &
  \begin{tabular}[c]{@{}c@{}}AUROC - Total\\ Uncertainty $\uparrow$ \end{tabular}  \\ \midrule

\multirow{2}{*}{SGLD} & STL10 &  0.677 & 0.684 \\
& SVHN & 0.948 & 0.945 \\

\multirow{2}{*}{SGHMC} & STL10 & 0.682 & 0.687 \\
& SVHN & 0.949 & 0.955 \\
                       
\multirow{2}{*}{cSGLD} & STL10 &0.624 & 0.641 \\
& SVHN & 0.966 & 0.968 \\

\multirow{2}{*}{cSGHMC} & STL10 & 0.631 & 0.657 \\
& SVHN & 0.920 & 0.945 \\
                       
\multirow{2}{*}{SWAG} & STL10 & 0.618 & 0.671 \\
& SVHN & 0.878 & 0.908 \\
                       
\multirow{2}{*}{PCA + ESS (SI)} & STL10 & 0.673 & 0.677 \\
& SVHN & 0.949 & 0.947 \\
 
\multirow{2}{*}{MC dropout} & STL10 & 0.665 & 0.695 \\
& SVHN & 0.926 & 0.938 \\

\multirow{2}{*}{SGD} & STL10 & N/A & 0.682 \\
& SVHN & N/A & 0.892 \\\bottomrule
\end{tabular}
\end{table*}

\begin{table*}[htbp]
  \caption{Comparision of Misclassification detection while using ResNet50 on CIFAR10.}
  \label{tab:misclass-resnet50-cifar10}
  \centering
\resizebox{\linewidth}{!}{
\begin{tabular}{@{}cccllll@{}}
\toprule
Inference &
  \begin{tabular}[c]{@{}c@{}}AUROC- Model\\ Uncertainty $\uparrow$ \end{tabular} &
  \begin{tabular}[c]{@{}c@{}}AUROC - Total\\ Uncertainty $\uparrow$ \end{tabular} &
  \begin{tabular}[c]{@{}l@{}}AUROC- Model\\ Confidence $\uparrow$ \end{tabular} &
  \multicolumn{1}{c}{\begin{tabular}[c]{@{}c@{}}AUCPR- Model\\ Uncertainty $\uparrow$ \end{tabular}} &
  \multicolumn{1}{c}{\begin{tabular}[c]{@{}c@{}}AUCPR - Total\\ Uncertainty $\uparrow$ \end{tabular}} &
  \begin{tabular}[c]{@{}l@{}}AUCPR- Model\\ Confidence $\uparrow$ \end{tabular} \\ \midrule
SGLD  & 0.945 & 0.949 & 0.950 & 0.422 & 0.468 & 0.480 \\ 
SGHMC  & 0.943 & 0.949 & 0.950 & 0.434 & 0.466 & 0.488 \\ 
cSGLD  & 0.927 & 0.943 & 0.946 & 0.321 & 0.355 & 0.382 \\ 
cSGHMC  & 0.885 & 0.935 & 0.943 & 0.311 & 0.390 & 0.444 \\ 
SWAG  & 0.890 & 0.927 & 0.927 & 0.418 & 0.479 & 0.472 \\ 
PCA + ESS (SI)  & 0.932 & 0.934 & 0.941 & 0.391 & 0.419 & 0.472 \\ 
MC dropout  & 0.946 & 0.947 & 0.947 & 0.455 & 0.485 & 0.477 \\ 
SGD  & N/A & 0.937 & 0.936 & N/A & 0.464 & 0.456 \\ \bottomrule
\end{tabular}
}
\end{table*}

%%%%%%%%%%%%%%%%%%%%%%%%%%%%%%%%%%%%%%%%%%%%%%%%%%
% RESNET 50: CIFAR100
%%%%%%%%%%%%%%%%%%%%%%%%%%%%%%%%%%%%%%%%%%%%%%%%%%

\begin{table*}[htbp]
  \caption{Comparison of predictive performance and decision making cost while using ResNet50 on CIFAR100.}
  \label{tab:pred-resnet50-cifar100}
  \centering
% \resizebox{\linewidth}{!}{
\begin{tabular}{@{}cccccc@{}}
\toprule
Inference & Accuracy $\uparrow$ & NLL $\downarrow$  & BS  $\downarrow$  & ECE $\downarrow$  & Decision Cost $\downarrow$ \\ \midrule
SGLD  & 0.751 & 1.079 & 0.364 & 0.107 & 277.500  \\ 
SGHMC  & 0.755 & 1.084 & 0.362 & 0.103 & 272.500  \\ 
cSGLD  & 0.804 & 0.711 & 0.272 & 0.019 & 211.900  \\ 
cSGHMC  & 0.814 & 0.667 & 0.261 & 0.012 & 198.500  \\ 
SWAG  & 0.735 & 1.221 & 0.400 & 0.135 & 295.200  \\ 
PCA + ESS (SI)  & 0.761 & 0.920 & 0.335 & 0.032 & 261.200  \\ 
MC dropout  & 0.786 & 1.006 & 0.330 & 0.115 & 233.100  \\ 
SGD  & 0.732 & 1.302 & 0.408 & 0.148 & 303.700  \\   \bottomrule
        %   &          &       &       &       &               \\ \bottomrule
\end{tabular}
% }
\end{table*}

\begin{table*}[htbp]
\caption{Comparison of OOD detection performance while using ResNet50 on CIFAR100.}
  \label{tab:ood-resnet50-cifar100}
  \centering
\begin{tabular}{@{}cccc@{}}
\toprule
Inference &
  \begin{tabular}[c]{@{}c@{}}OOD\\ Dataset\end{tabular} &
  \begin{tabular}[c]{@{}c@{}}AUROC- Model\\ Uncertainty $\uparrow$ \end{tabular}  &
  \begin{tabular}[c]{@{}c@{}}AUROC - Total\\ Uncertainty $\uparrow$ \end{tabular} \\ \midrule

\multirow{2}{*}{SGLD} & STL10 &  0.769 & 0.782 \\
& SVHN & 0.772 & 0.802 \\

\multirow{2}{*}{SGHMC} & STL10 & 0.773 & 0.784 \\
& SVHN & 0.809 & 0.823 \\
                       
\multirow{2}{*}{cSGLD} & STL10 & 0.806 & 0.827 \\
& SVHN & 0.809 & 0.816 \\

\multirow{2}{*}{cSGHMC} & STL10 & 0.804 & 0.832 \\
& SVHN & 0.791 & 0.823 \\
                       
\multirow{2}{*}{SWAG} & STL10 & 0.748 & 0.778 \\
& SVHN & 0.732 & 0.771 \\
                       
\multirow{2}{*}{PCA + ESS (SI)} & STL10 & 0.779 & 0.797 \\
& SVHN & 0.816 & 0.807 \\

\multirow{2}{*}{MC dropout} & STL10 &0.785 & 0.801 \\
& SVHN & 0.755 & 0.752 \\ 

\multirow{2}{*}{SGD} & STL10 & N/A & 0.765 \\
& SVHN & N/A & 0.763 \\

\bottomrule
\end{tabular}
\end{table*}

\begin{table*}[htbp]
  \caption{Comparision of Misclassification detection while using ResNet50 on CIFAR100.}
  \label{tab:misclass-resnet50-cifar100}
  \centering
\resizebox{\linewidth}{!}{
\begin{tabular}{@{}cccllll@{}}
\toprule
Inference &
  \begin{tabular}[c]{@{}c@{}}AUROC- Model\\ Uncertainty $\uparrow$ \end{tabular} &
  \begin{tabular}[c]{@{}c@{}}AUROC - Total\\ Uncertainty $\uparrow$ \end{tabular} &
  \begin{tabular}[c]{@{}l@{}}AUROC- Model\\ Confidence $\uparrow$ \end{tabular} &
  \multicolumn{1}{c}{\begin{tabular}[c]{@{}c@{}}AUCPR- Model\\ Uncertainty $\uparrow$ \end{tabular}} &
  \multicolumn{1}{c}{\begin{tabular}[c]{@{}c@{}}AUCPR - Total\\ Uncertainty $\uparrow$ \end{tabular}} &
  \begin{tabular}[c]{@{}l@{}}AUCPR- Model\\ Confidence $\uparrow$ \end{tabular} \\ \midrule
SGLD  & 0.870 & 0.882 & 0.879 & 0.648 & 0.683 & 0.672 \\ 
SGHMC  & 0.863 & 0.873 & 0.871 & 0.635 & 0.659 & 0.651 \\ 
cSGLD  & 0.872 & 0.880 & 0.891 & 0.564 & 0.623 & 0.654 \\ 
cSGHMC  & 0.873 & 0.879 & 0.890 & 0.572 & 0.593 & 0.626 \\ 
SWAG  & 0.855 & 0.870 & 0.869 & 0.625 & 0.671 & 0.667 \\ 
PCA + ESS (SI)  & 0.858 & 0.863 & 0.877 & 0.618 & 0.634 & 0.667 \\ 
MC dropout  & 0.875 & 0.880 & 0.877 & 0.613 & 0.639 & 0.624 \\ 
SGD  & N/A & 0.873 & 0.870 & N/A & 0.687 & 0.680 \\   \bottomrule
\end{tabular}
}
\end{table*}

%%%%%%%%%%%%%%%%%%%%%%%%%%%%%%%%%%%%%%%%%%%%%%%%%%
% WideResNet28x10: CIFAR10
%%%%%%%%%%%%%%%%%%%%%%%%%%%%%%%%%%%%%%%%%%%%%%%%%%

\begin{table*}[htbp]
  \caption{Comparison of predictive performance and decision making cost while using WideResNet28x10 on CIFAR10.}
  \label{tab:pred-wide-cifar10}
  \centering
% \resizebox{\linewidth}{!}{
\begin{tabular}{@{}cccccc@{}}
\toprule
Inference & Accuracy $\uparrow$ & NLL $\downarrow$  & BS  $\downarrow$  & ECE $\downarrow$  & Decision Cost $\downarrow$ \\ \midrule
SGLD  & 0.965 & 0.113 & 0.054 & 0.004 & 115.800  \\ 
SGHMC  & 0.965 & 0.114 & 0.053 & 0.004 & 112.200  \\ 
cSGLD  & 0.967 & 0.104 & 0.050 & 0.006 & 102.900  \\ 
cSGHMC  & 0.957 & 0.196 & 0.072 & 0.079 & 140.800  \\ 
SWAG  & 0.919 & 0.260 & 0.121 & 0.028 & 270.000  \\ 
PCA + ESS (SI)  & 0.951 & 0.177 & 0.082 & 0.054 & 163.100  \\ 
MC dropout  & 0.957 & 0.158 & 0.067 & 0.019 & 149.000  \\ 
SGD  & 0.963 & 0.138 & 0.060 & 0.018 & 117.100  \\     \bottomrule
        %   &          &       &       &       &               \\ \bottomrule
\end{tabular}
% }
\end{table*}

\begin{table*}[htbp]
\caption{Comparison of OOD detection performance while using WideResNet28x10 on CIFAR10.}
  \label{tab:ood-wide-cifar10}
  \centering
\begin{tabular}{@{}cccc@{}}
\toprule
Inference &
  \begin{tabular}[c]{@{}c@{}}OOD\\ Dataset\end{tabular} &
  \begin{tabular}[c]{@{}c@{}}AUROC- Model\\ Uncertainty $\uparrow$ \end{tabular}  &
  \begin{tabular}[c]{@{}c@{}}AUROC - Total\\ Uncertainty $\uparrow$ \end{tabular}  \\ \midrule

\multirow{2}{*}{SGLD} & STL10 &  0.680 & 0.680 \\
& SVHN & 0.951 & 0.963 \\

\multirow{2}{*}{SGHMC} & STL10 & 0.678 & 0.683 \\
& SVHN & 0.956 & 0.967 \\
                       
\multirow{2}{*}{cSGLD} & STL10 & 0.685 & 0.686 \\
& SVHN & 0.968 & 0.974 \\

\multirow{2}{*}{cSGHMC} & STL10 &0.614 & 0.648 \\
& SVHN & 0.864 & 0.952 \\

\multirow{2}{*}{SWAG} & STL10 & 0.649 & 0.667 \\
& SVHN & 0.914 & 0.943 \\
                       
\multirow{2}{*}{PCA + ESS (SI)} & STL10 &0.663 & 0.673 \\
& SVHN & 0.897 & 0.970 \\

\multirow{2}{*}{MC dropout} & STL10 &0.672 & 0.688 \\
& SVHN & 0.897 & 0.922 \\

\multirow{2}{*}{SGD} & STL10 & N/A & 0.667 \\
& SVHN & N/A & 0.963 \\
\bottomrule
\end{tabular}
\end{table*}

\begin{table*}[htbp]
  \caption{Comparision of Misclassification detection while using WideResNet28x10 on CIFAR10.}
  \label{tab:misclass-wide-cifar10}
  \centering
\resizebox{\linewidth}{!}{
\begin{tabular}{@{}cccllll@{}}
\toprule
Inference &
  \begin{tabular}[c]{@{}c@{}}AUROC- Model\\ Uncertainty $\uparrow$ \end{tabular} &
  \begin{tabular}[c]{@{}c@{}}AUROC - Total\\ Uncertainty $\uparrow$ \end{tabular} &
  \begin{tabular}[c]{@{}l@{}}AUROC- Model\\ Confidence $\uparrow$ \end{tabular} &
  \multicolumn{1}{c}{\begin{tabular}[c]{@{}c@{}}AUCPR- Model\\ Uncertainty $\uparrow$ \end{tabular}} &
  \multicolumn{1}{c}{\begin{tabular}[c]{@{}c@{}}AUCPR - Total\\ Uncertainty $\uparrow$ \end{tabular}} &
  \begin{tabular}[c]{@{}l@{}}AUCPR- Model\\ Confidence $\uparrow$ \end{tabular}\\ \midrule
SGLD  & 0.952 & 0.954 & 0.955 & 0.402 & 0.414 & 0.435 \\ 
SGHMC  & 0.954 & 0.956 & 0.958 & 0.380 & 0.415 & 0.439 \\ 
cSGLD  & 0.949 & 0.952 & 0.953 & 0.354 & 0.383 & 0.406 \\ 
cSGHMC  & 0.889 & 0.936 & 0.945 & 0.298 & 0.382 & 0.452 \\ 
SWAG  & 0.900 & 0.915 & 0.916 & 0.375 & 0.468 & 0.468 \\ 
PCA + ESS (SI)  & 0.918 & 0.931 & 0.948 & 0.337 & 0.387 & 0.478 \\ 
MC dropout  & 0.946 & 0.947 & 0.947 & 0.432 & 0.467 & 0.467 \\ 
SGD  & N/A & 0.941 & 0.942 & N/A & 0.390 & 0.387 \\  \bottomrule
\end{tabular}
}
\end{table*}

%%%%%%%%%%%%%%%%%%%%%%%%%%%%%%%%%%%%%%%%%%%%%%%%%%
% WideResNet28x10: CIFAR100
%%%%%%%%%%%%%%%%%%%%%%%%%%%%%%%%%%%%%%%%%%%%%%%%%%

\begin{table*}[htbp]
  \caption{Comparison of predictive performance and decision making cost while using WideResNet28x10 on CIFAR100.}
  \label{tab:pred-wide-cifar100}
  \centering
% \resizebox{\linewidth}{!}{
\begin{tabular}{@{}cccccc@{}}
\toprule
Inference & Accuracy $\uparrow$ & NLL $\downarrow$  & BS  $\downarrow$  & ECE $\downarrow$  & Decision Cost $\downarrow$ \\ \midrule
SGLD  & 0.809 & 0.760 & 0.278 & 0.066 & 204.300  \\ 
SGHMC  & 0.798 & 0.815 & 0.292 & 0.076 & 216.200  \\ 
cSGLD  & 0.832 & 0.640 & 0.242 & 0.033 & 179.300  \\ 
cSGHMC  & 0.821 & 0.666 & 0.258 & 0.059 & 191.800  \\ 
SWAG  & 0.710 & 1.149 & 0.414 & 0.110 & 316.900  \\ 
PCA + ESS (SI)  & 0.817 & 0.656 & 0.263 & 0.038 & 196.600  \\ 
MC dropout  & 0.798 & 0.846 & 0.293 & 0.081 & 214.300  \\ 
SGD  & 0.806 & 0.785 & 0.280 & 0.046 & 205.100  \\    \bottomrule
\end{tabular}
% }
\end{table*}

\begin{table*}[htbp]
\caption{Comparison of OOD detection performance while using WideResNet28x10 on CIFAR100.}
  \label{tab:ood-wide-cifar100}
  \centering
\begin{tabular}{@{}cccc@{}}
\toprule
Inference &
  \begin{tabular}[c]{@{}c@{}}OOD\\ Dataset\end{tabular} &
  \begin{tabular}[c]{@{}c@{}}AUROC- Model\\ Uncertainty $\uparrow$ \end{tabular}  &
  \begin{tabular}[c]{@{}c@{}}AUROC - Total\\ Uncertainty $\uparrow$ \end{tabular}  \\ \midrule

\multirow{2}{*}{SGLD} & STL10 &  0.797 & 0.822 \\
& SVHN & 0.784 & 0.791 \\

\multirow{2}{*}{SGHMC} & STL10 & 0.799 & 0.814 \\
& SVHN & 0.768 & 0.794 \\
                       
\multirow{2}{*}{cSGLD} & STL10 & 0.791 & 0.846 \\
& SVHN & 0.767 & 0.782 \\

\multirow{2}{*}{cSGHMC} & STL10 & 0.816 & 0.845 \\
& SVHN & 0.786 & 0.837 \\
                       
\multirow{2}{*}{SWAG} & STL10 &0.732 & 0.753 \\
& SVHN & 0.672 & 0.704 \\
                       
\multirow{2}{*}{PCA + ESS (SI)} & STL10 &0.813 & 0.827 \\
& SVHN & 0.760 & 0.814 \\

\multirow{2}{*}{MC dropout} & STL10 &0.798 & 0.815 \\
& SVHN & 0.642 & 0.645 \\ 

\multirow{2}{*}{SGD} & STL10 & N/A & 0.820 \\
& SVHN & N/A & 0.732 \\ \bottomrule
\end{tabular}

\end{table*}

\begin{table*}[htbp]
  \caption{Comparision of Misclassification detection while using WideResNet28x10 on CIFAR100.}
  \label{tab:misclass-wide-cifar100}
  \centering
\resizebox{\linewidth}{!}{
\begin{tabular}{@{}cccllll@{}}
\toprule
Inference &
  \begin{tabular}[c]{@{}c@{}}AUROC- Model\\ Uncertainty $\uparrow$ \end{tabular} &
  \begin{tabular}[c]{@{}c@{}}AUROC - Total\\ Uncertainty $\uparrow$ \end{tabular} &
  \begin{tabular}[c]{@{}l@{}}AUROC- Model\\ Confidence $\uparrow$ \end{tabular} &
  \multicolumn{1}{c}{\begin{tabular}[c]{@{}c@{}}AUCPR- Model\\ Uncertainty $\uparrow$ \end{tabular}} &
  \multicolumn{1}{c}{\begin{tabular}[c]{@{}c@{}}AUCPR - Total\\ Uncertainty $\uparrow$ \end{tabular}} &
  \begin{tabular}[c]{@{}l@{}}AUCPR- Model\\ Confidence $\uparrow$ \end{tabular} \\ \midrule
SGLD  & 0.881 & 0.888 & 0.892 & 0.579 & 0.616 & 0.629 \\ 
SGHMC  & 0.884 & 0.893 & 0.894 & 0.625 & 0.650 & 0.654 \\ 
cSGLD  & 0.870 & 0.874 & 0.892 & 0.499 & 0.543 & 0.595 \\ 
cSGHMC  & 0.854 & 0.865 & 0.888 & 0.516 & 0.553 & 0.609 \\ 
SWAG  & 0.837 & 0.857 & 0.860 & 0.601 & 0.675 & 0.686 \\ 
PCA + ESS (SI)  & 0.853 & 0.868 & 0.888 & 0.520 & 0.559 & 0.603 \\ 
MC dropout  & 0.883 & 0.887 & 0.887 & 0.617 & 0.638 & 0.637 \\ 
SGD  & N/A & 0.869 & 0.879 & N/A & 0.586 & 0.622 \\   \bottomrule
\end{tabular}
}
\end{table*}
\balance

\end{document}